%% file: neurips_2019_index.tex
\documentclass{article}
\input{math_commands.tex}

     \PassOptionsToPackage{numbers, compress}{natbib}
    \usepackage[preprint]{neurips_2019}
\usepackage[utf8]{inputenc} % allow utf-8 input
\usepackage[T1]{fontenc}    % use 8-bit T1 fonts
\usepackage{hyperref}       % hyperlinks
\usepackage{url}            % simple URL typesetting
\usepackage{booktabs}       % professional-quality tables
\usepackage{amsfonts}       % blackboard math symbols
\usepackage{nicefrac}       % compact symbols for 1/2, etc.
\usepackage{microtype}      % microtypography
\usepackage[titlenumbered,ruled]{algorithm2e}
\usepackage{algorithmic}
\usepackage{threeparttable}
\usepackage{makecell}
\usepackage{multirow}
\usepackage{multicol}
\usepackage{amsmath}
\usepackage{graphicx}

\title{Network Pruning for Low-Rank Binary Indexing}
\author{Dongsoo Lee\textsuperscript{1}, Se Jung Kwon\textsuperscript{1}, Byeongwook Kim\textsuperscript{1}, Parichay Kapoor\textsuperscript{1}, Gu-Yeon Wei\textsuperscript{1,2} \\
  \textsuperscript{1} Samsung Research, Seoul, Republic of Korea \\
  \textsuperscript{2} Harvard University, Cambridge, MA \\
  \texttt{\{dongsoo3.lee, sejung0.kwon, byeonguk.kim, pk.kapoor, gy.wei\}@samsung.com}\\
}

\begin{document}

\maketitle

\begin{abstract}
  Pruning is an efficient model compression technique to remove redundancy in the connectivity of deep neural networks (DNNs).
  Computations using sparse matrices obtained by pruning parameters, however, exhibit vastly different parallelism depending on the index representation scheme.
  As a result, fine-grained pruning has not gained much attention due to its irregular index form leading to large memory footprint and low parallelism for convolutions and matrix multiplications.
  In this paper, we propose a new network pruning technique that generates a low-rank binary index matrix to compress index data while decompressing index data is performed by simple binary matrix multiplication.
  This proposed compression method finds a particular fine-grained pruning mask that can be decomposed into two binary matrices.
  We also propose a tile-based factorization technique that not only lowers memory requirements but also enhances compression ratio.
  Various DNN models can be pruned with much fewer indexes compared to previous sparse matrix formats while maintaining the same pruning rate.
\end{abstract}

\section{Introduction}

Numerous parameter pruning techniques have been introduced based on the observation that significant amounts of connectivity can be removed without sacrificing model accuracy.
Current active research strives to enhance pruning rate at the cost of additional computations \cite{DNS, sparseVD}, to reduce computational complexity of pruning procedures \cite{suyog_prune, SHan_2015}, and to find the underlying principles of pruning \cite{rethinking, lottery}, to name a few.

Unfortunately, conventional pruning techniques present serious implementation challenges when deployed in highly parallel computing systems due to their sparse matrix formats.
If pruning is performed in a fine-grained manner (i.e., each parameter is evaluated to be pruned) to improve overall pruning rate, then each row/column or block exhibits different sparsity.
Consequently, row/column-wise or block-wise computations require vastly different computation latency leading to significantly degraded parallelism \cite{EIE}.
On the other hand, successful DNN deployment depends on accelerating matrix multiplications and convolutions using thread-oriented computational resources such as GPUs and multi-threaded CPUs \cite{large_scale_distributed}.
Consequently, fine-grained pruning techniques have not been adopted by industry.
Note that a few attempts have been made to utilize sparsity to expedite inference for batch size of 1 \cite{EIE}.
However, if parameter reuse rate increases due to convolutions or a large batch size, then sparse matrices show even slower performance than dense matrices before pruning \cite{EIE, cambricon-x}.
Recent pruning techniques suggest removing connectivity in a well-structured form, \cite{channel_pruning_2, channel_pruning_1}, or in a block-level \cite{scalpel2017, block-sparse}.

\begin{figure}[t]
\begin{center}
\includegraphics[width=0.85\linewidth]{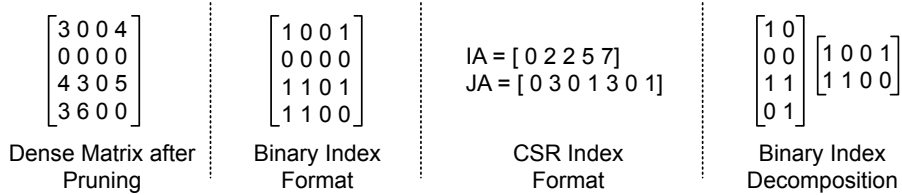}
\caption{Comparison on various sparse matrix representation formats.}
\label{fig:format_comp}
\end{center}
\end{figure}

In this paper, we propose a new fine-grained pruning method to find an efficient sparse matrix representation based on {\it binary index-matrix factorization}.
Figure~\ref{fig:format_comp} shows a dense matrix after pruning redundant parameters and various masking index representations.
The binary index form is a regular structure that can utilize full memory bandwidth even though sparsity does not reduce memory footprint.
Compressed sparse row (CSR) indexing presents an irregular structure that reduces storage requirements according to sparsity.
In contrast, our proposed index compression scheme decomposes a binary index matrix into two small binary matrices in order to maintain regular structure.
Binary matrix multiplication is inherently parallelizable and sparse quantized networks can be decoded efficiently using recent approaches \cite{viterbi_quantized}.
In order to accomplish this indexing form, we propose an algorithm that finds a particular fine-grained pruning result by generating a low-rank binary index matrix.
Since factorization may not exactly reproduce the original binary index matrix, we investigate whether a low-rank binary index matrix produces a pruning result while maintaining allowable model accuracy.
Our proposed binary matrix factorization technique significantly reduces the amount of index data compared to using CSR, as demonstrated in the next sections.
We also introduce a tiling technique to alleviate the burden of binary matrix multiplication and further improve compression ratio.

Recently, a regular-structured index compression method has been proposed using the Viterbi algorithm \cite{lee2018viterbibased}, which explores sequences to be used as pruning mask bits, decompressed by Viterbi encoders.
Even though compression rate improves over CSR, for every one input bit, a large number of XOR gates and delay units is required. In comparison, our proposed decompression relies on simple binary matrix multiplications and achieves even higher compression.

\section{Binary Matrix Factorization for Pruning Index}

Suppose that a $5 \times 5$ weight matrix $\displaystyle \mW$ is given as
\begin{equation}
\label{eq:1}
{\displaystyle \mW}=
\begin{bmatrix}
-0.1 & 0.9 & 1.2 & -0.2 & -0.6 \\
1.8 & 0.2 & -0.7 & -1.6 & 0.6 \\
-0.1 & -1.7 & 0.1 & -0.3 & 1.2 \\
-0.4 & 1.4 & -0.9 & 0.6 & 1.4 \\
-1.1 & 0.5 & 1.0 & 1.0 & -0.3
\end{bmatrix}.
\end{equation}
Following the magnitude-based pruning method \cite{SHan_2015}, all weights with magnitude smaller than a certain threshold value are pruned to zero.
For example, for a threshold of 0.7, we obtain the following pruning index (i.e., binary masking layer) matrix $\displaystyle \mI$ as
\begin{equation}
\label{eq:2}
{\displaystyle \mI}=
\begin{bmatrix}
0 & 1 & 1 & 0 & 0 \\
1 & 0 & 1 & 1 & 0 \\
0 & 1 & 0 & 0 & 1 \\
0 & 1 & 1 & 0 & 1 \\
1 & 0 & 1 & 1 & 0
\end{bmatrix}.
\end{equation}
It is important to note that magnitude-based pruning is not an optimal solution \cite{lee2018viterbibased, optimalbrain, deepcompression} to maximize the pruning rate, i.e., a variety of masking layers exist for the same pruning rate.

Binary matrix factorization (BMF) \cite{BMF} is a reasonable approach to compress $\displaystyle \mI$.
For $\mI\in\{0,1\}^{m\times n}$, BMF tries to find the best $\displaystyle \mI_p^{m \times k}$ and $\displaystyle \mI_z^{k \times n}$ to approximate $\displaystyle \mI$ as $\displaystyle \mI$ $\approx$ $\displaystyle \mI_p$ $\otimes$ $\displaystyle \mI_z$ $=\displaystyle \mI_a$, where $k$ corresponds to the binary matrix factorization rank and $\otimes$ stands for binary matrix multiplication.
A binary product of $\displaystyle \mI_p$ and $\displaystyle \mI_z$ is then defined as
\begin{equation}
\label{eq:3}
    (\emI _a)_{i,j} = \bigvee_{l=1}^{k} (\emI _p)_{i,l} \wedge (\emI _z)_{l,j}.
\end{equation}
While BMF should minimize the number of mismatched bits between $\displaystyle \mI$ and $\displaystyle \mI_a$,
such an optimization is NP-hard. Hence, several heuristic methods for BMF have been proposed in the literature \cite{BMF}.
Moreover, BMF using $\displaystyle \mI$ (without referring to $\displaystyle \mW$) lacks weight magnitude information.
Yet, in the context of magnitude-based pruning, weights of small magnitude should be pruned with higher probability (i.e., lower importance).
Therefore, we explore a method that preserves the importance of each weight when it is not possible to exactly reproduce $\displaystyle \mI$ through matrix factorization.

\subsection{Binary Matrix Factorization based on Non-Negative Matrix Factorization}

Non-negative matrix factorization (NMF) factorizes a real-valued matrix $\displaystyle \mH$ into two real-valued matrices $\displaystyle \mH_1$ and $\displaystyle \mH_2$ under the constraint that all three matrices consist of non-negative elements.
Similar to singular-value decomposition (SVD), NMF attempts to minimize $\displaystyle || \mH - \mH_1\mH_2 || ^2_{\mathcal{F}}$,
where $\displaystyle || \mH ||_{\mathcal{F}}$ denotes the Frobenius norm of the matrix $\displaystyle \mH$.
The property of non-negativity in NMF is useful for analysis of various multivariate data. Numerous numerical approximations of NMF have been suggested since an exact solution of NMF is not generally available \cite{NMF, BMF}.

In our proposed technique, we take the magnitude of each element of $\displaystyle \mW$ to generate $\displaystyle \mM$ (i.e., $\displaystyle \emM_{i,j} = | \emW_{i,j} |$). and $\displaystyle \mM$ is factorized by an NMF library (e.g., \cite{Zitnik2012}) into two matrices $\displaystyle \mM_p$ and $\displaystyle \mM_z$.
For example, a magnitude matrix $\displaystyle \mM$ of the matrix $\displaystyle \mW$ of Eq. (\ref{eq:1}) can be factorized into
\begin{equation}
\label{eq:4}
\displaystyle \mM_p = \begin{bmatrix}
0.2 & 0.5 \\
1.3 & 0.0 \\
0.0 & 0.9 \\
0.3 & 0.8 \\
0.8 & 0.2 
\end{bmatrix}
, \quad \displaystyle \mM_z = \begin{bmatrix}
1.3 & 0.1 & 0.7 & 1.2 & 0.3 \\
0.0 & 1.8 & 0.7 & 0.2 & 1.3
\end{bmatrix},
\end{equation}
where rank $k$ is 2.

The next step is to convert $\displaystyle \mM_p$ and $\displaystyle \mM_z$ into two binary matrices $\displaystyle \mI_p$ and $\displaystyle \mI_z$ using threshold values $T_p$ and $T_z$ (i.e., $\displaystyle (\emI_p)_{i,j}$ $=1$ if $\displaystyle (\emM_p)_{i,j} \geq T_p$, or 0 otherwise).
The sparsity of $\displaystyle \mI_p$ and $\displaystyle \mI_z$ can each be controlled by $T_p$ and $T_z$. Our goal is to achieve similar sparsity between $\displaystyle \mI$ and $\displaystyle \mI_p \otimes \mI_z$.
Suppose that $T_p=0.5$ and $T_z=0.6$ are carefully chosen to produce similar sparsity as $\displaystyle \mI$. We obtain
\begin{equation}
\label{eq:5}
\displaystyle \mI_p = \begin{bmatrix}

0 & 1 \\
1 & 0 \\
0 & 1 \\
0 & 1 \\
1 & 0
\end{bmatrix}
, \quad \mI_z = \begin{bmatrix}
1 & 0 & 1 & 1 & 0 \\
0 & 1 & 1 & 0 & 1
\end{bmatrix}
\end{equation}
and the binary product of $\displaystyle \mI_p$ and $\displaystyle \mI_z$ becomes
\begin{equation}
\label{eq:6}
\displaystyle \mI_a=\mI_p \otimes \mI_z=
\begin{bmatrix}
0 & 1 & 1 & 0 & \underline{1} \\
1 & 0 & 1 & 1 & 0 \\
0 & 1 & \underline{1} & 0 & 1 \\
0 & 1 & 1 & 0 & 1 \\
1 & 0 & 1 & 1 & 0 
\end{bmatrix}.
\end{equation}
Compared with the pruning-index matrix $\displaystyle \mI$ in Eq. (\ref{eq:2}), there are 2 mismatched elements (underlined in Eq. (\ref{eq:6})).

The rationale behind this approach is as follows: 1) If $\displaystyle \emM_{i,j}$ is large, then its corresponding $k$ components of $\displaystyle \mM_p$ and $\displaystyle \mM_z$ (i.e., $\displaystyle ({\mM_p})_{i, :}$ and $\displaystyle ({\mM_z})_{:, j}$) will also be large with high probability and, correspondingly, 2) Binary matrix conversion using $T_p$ and $T_z$ would yield a high probability of being `1' within $\displaystyle \mI_p$ and $\displaystyle \mI_z$ if the corresponding $\displaystyle \emM_{i,j}$ is large.
Let $S$, $S_a$, $S_p$, and $S_z$ be the sparsity of $\displaystyle \mI$, $\displaystyle \mI_a$, $\displaystyle \mI_p$, and $\displaystyle \mI_z$, respectively.
From the dot product operation in Eq. (\ref{eq:6}), the expression for pruning rate $S$ becomes
\begin{equation}
\label{eq:7}
    S = (( 1-(1-S_p)(1-S_z))^{k},
\end{equation}
assuming that the probability of a bit being `0' in $\displaystyle \mI_p$ and $\displaystyle \mI_z$ follows $S_p$ and $S_z$.
Then, $S_z$ = $(S^{1/k} - S_p) / (1-S_p)$, which needs to be fine-tuned in practice.
If $T_p$ and associated $S_p$ are given, then $T_z$ and $S_z$ are automatically determined by the target pruning rate.
Subsequently, given $\displaystyle \mW$ and rank $k$, it is necessary to find a certain $S_p$ that produces the best masking layer for pruning.

\begin{algorithm}[t]
   \SetAlgoLined
   \SetKwInOut{Input}{input}
   \SetKwInOut{Output}{output}
   \caption{Binary pruning-index-data matrix factorization}
   \label{alg:nmf}
   \Input{$\displaystyle \mW\in\R^{m\times n}$, rank $k$, target sparsity $S$}
   \Output{$\displaystyle \mI_p\in\{0,1\}^{m\times k}$, $\displaystyle \mI_z\in\{0,1\}^{k \times n}$}
\begin{algorithmic}[1]
   \STATE Generate the magnitude matrix $\displaystyle \mM$ using $\displaystyle \mW$
   \STATE $\displaystyle \mM_p$, $\displaystyle \mM_z$ = NMF($\displaystyle \mM$, $k$)
   \STATE $Cost^{min} \leftarrow \infty$, \, $S_p^{min} \leftarrow 0.0$, \, $S_z^{min} \leftarrow 0.0$
   
   \FOR{$S_p=0.0$ {\bfseries to} $1.0$}
    \STATE Compute $S_z$ using Eq. (\ref{eq:7})
    \REPEAT 
    \STATE Convert $(\displaystyle \mM_p$,$\displaystyle \mM_z)$ into $(\displaystyle \mI_p$,$\displaystyle \mI_z)$ w/ $(S_p, S_z)$
    \STATE Adjust $S_z$ depending on ($S_a - S$)
    \UNTIL{$S_a \approx S$}
    \STATE $Cost \leftarrow \sum\limits_{\emI_{i,j}=1,(\emI_a)_{i,j}=0} \emM_{i,j}$ 
    \IF{$Cost^{min} > Cost$}
        \STATE $Cost^{min} \leftarrow Cost$, \, $S_p^{min} \leftarrow S_p$, \, $S_z^{min} \leftarrow S_z$
    \ENDIF
   \ENDFOR
   \STATE Convert $(\displaystyle \mM_p , \displaystyle \mM_z)$ into $(\displaystyle \mI_p , \displaystyle \mI_z)$ w/ $(S_p^{min} , S_z^{min})$
   \STATE \textbf{Return} $\displaystyle \mI_p$, $\displaystyle \mI_z$
\end{algorithmic}
\end{algorithm}

In order to optimize $S_p$, we define the cost function for pruning-index compression to be $\sum{\displaystyle \emM_{i,j}}$, where $\displaystyle (\emI)_{i,j}=1$ and $\displaystyle (\emI_a)_{i,j}=0$ (i.e., a sum of all unintentionally pruned weights' magnitude by binary matrix decomposition).
Algorithm~\ref{alg:nmf} describes a method to find the best $S_p$ by sweeping $S_p$ and monitoring $Cost$. Given the linear relationship between $S_a$ and $S_z$, the algorithm can use binary search to expedite adjustment of $S_z$.

\subsection{MNIST Case Study}

\begin{figure*}[t]
	\begin{minipage}{.320\textwidth}
		\includegraphics[width=1\linewidth]{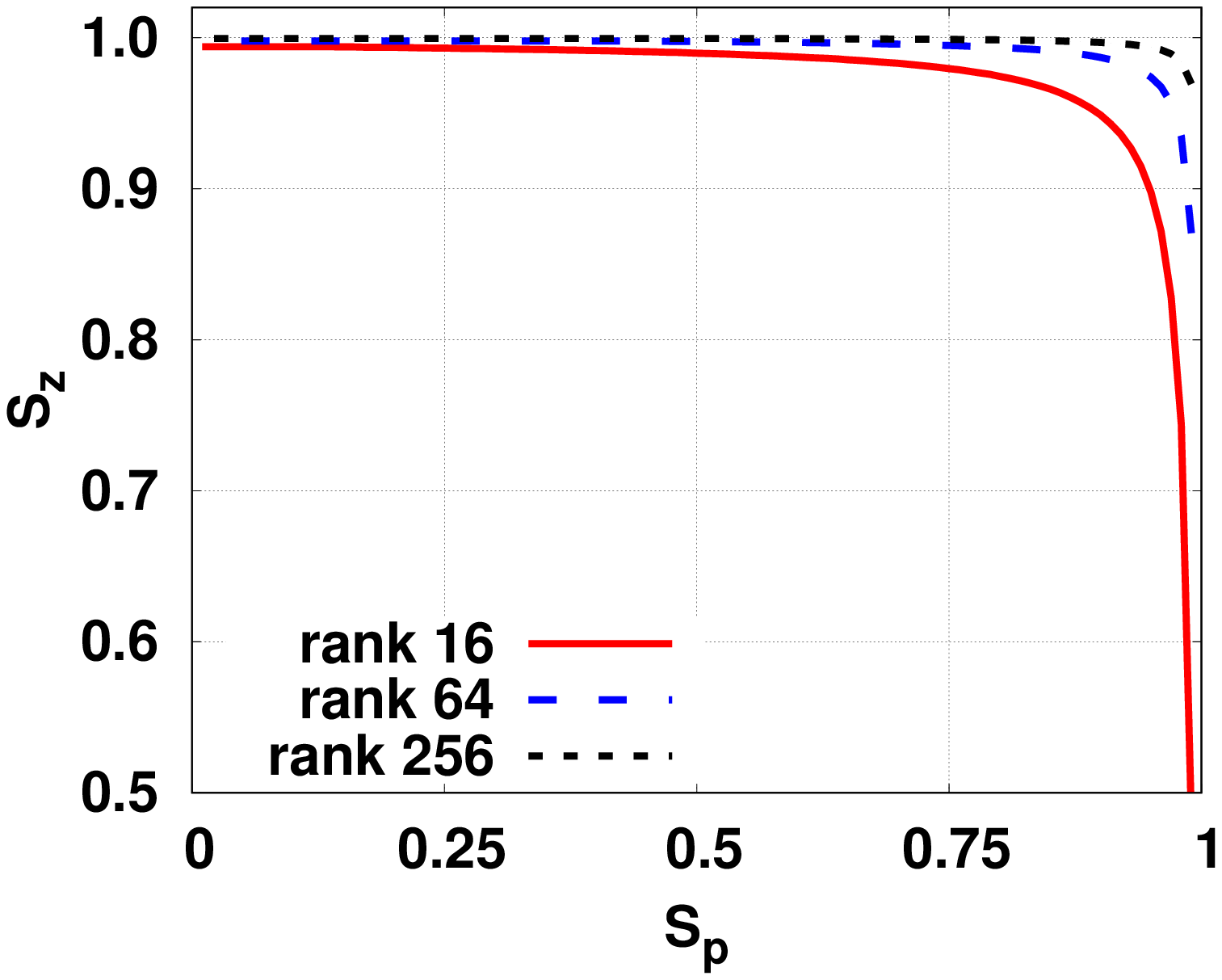}
	\end{minipage}
	\hfill
	\begin{minipage}{.320\textwidth}
		\includegraphics[width=1\linewidth]{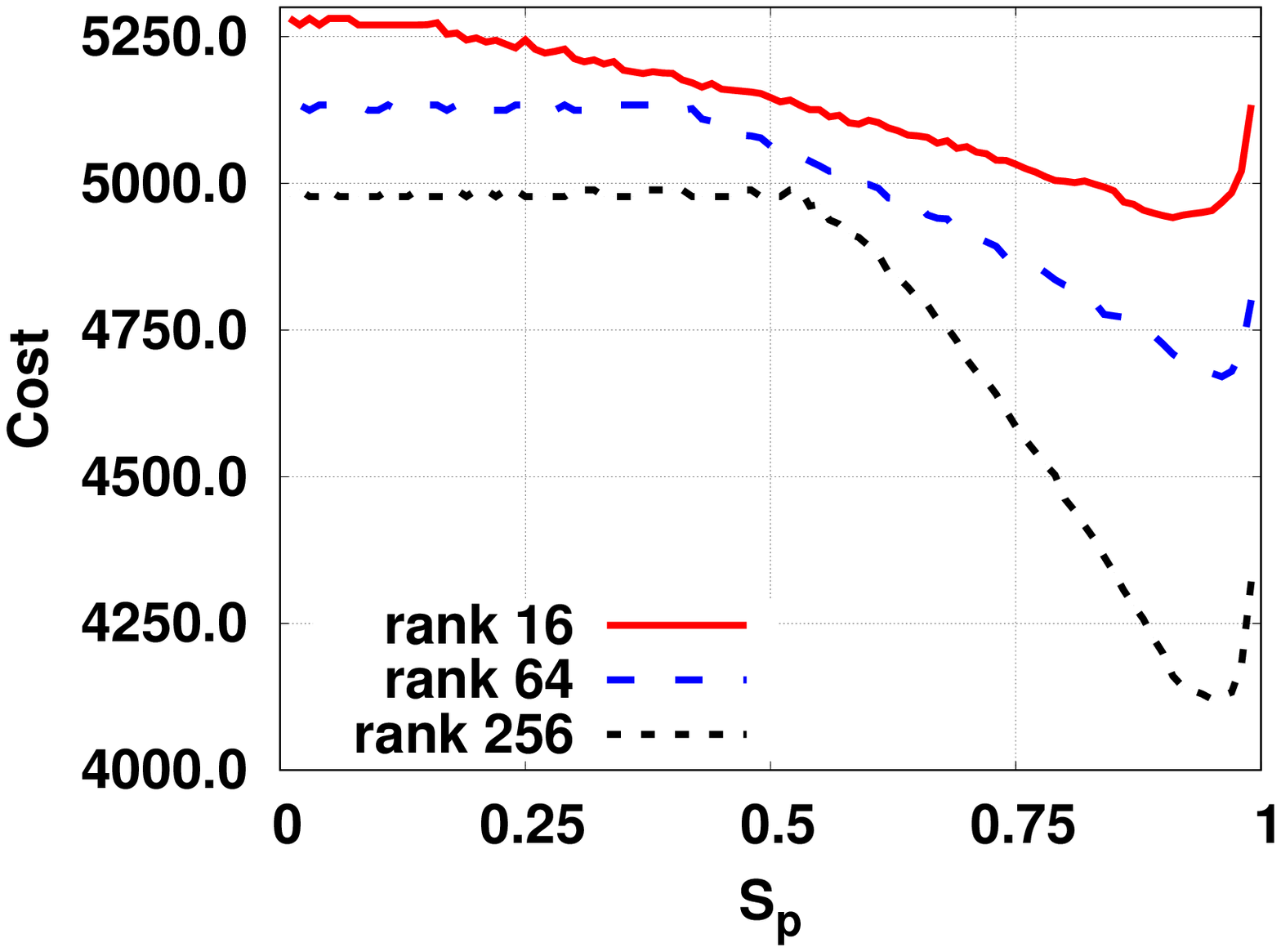}
	\end{minipage}
	\hfill
	\begin{minipage}{.320\textwidth}
		\includegraphics[width=1\linewidth]{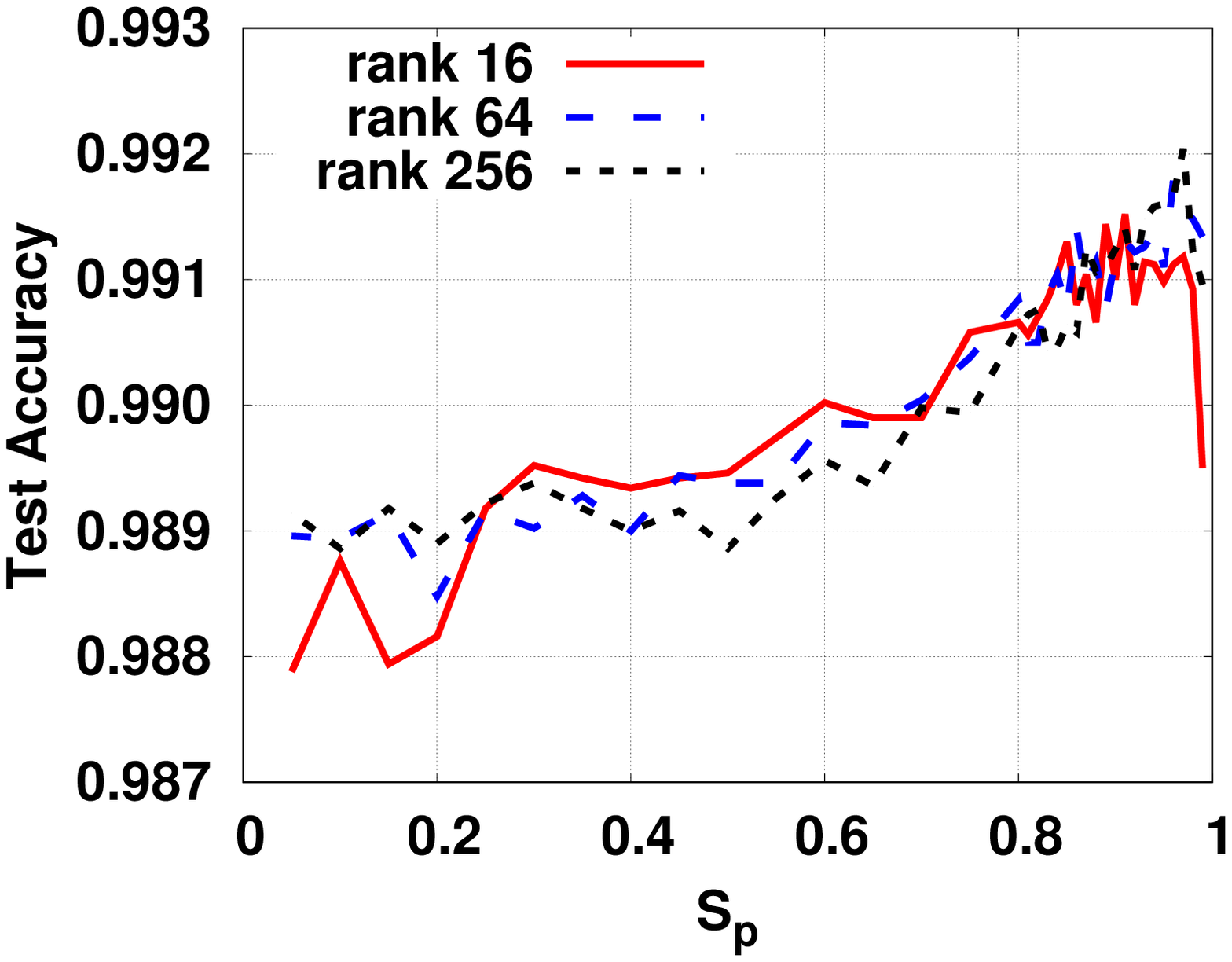}
	\end{minipage}
	\caption{$S_z$, $Cost$, and test accuracy for various $S_p$ when the FC1 layer's pruning-index data of LeNet-5 model is factorized by Algorithm \ref{alg:nmf} using $S$=0.95, $k$ = 16, 64, or 256.}
	\label{fig:sp_optimization}
\end{figure*}

We applied our proposed pruning-index compression technique to the LeNet-5 model with the MNIST dataset.
LeNet-5 consists of two convolutional layers and two fully-connected layers \cite{SHan_2015, DeepTwist}.
Since the FC1 layer dominates the memory footprint (93\%), we focus on the FC1 layer to investigate the effectiveness of our proposed method, but all layers are pruned with the same rates as in \cite{SHan_2015}.
FC1 involves a $(800\times500)$ weight matrix, where full rank is 500.
Figure~\ref{fig:sp_optimization} plots $S_z$, $Cost$, and test accuracy across a range of $S_p$ values.
Pruning rate $S$ for FC1 is 95\% and rank $k$ is 16, 64, or 256.
As $k$ increases, both $\displaystyle \mI_p$ and $\displaystyle \mI_z$ become more sparse and both $Cost$ and test accuracy improve.

After pre-training for 20K iterations, pruning via BMF is performed as described in Algorithm~\ref{alg:nmf}.
Then, retraining runs until the 60K\textsuperscript{th} iteration using a masking layer $\displaystyle \mI_a$.
The test accuracy (the average of 20 runs) is measured at the 20K\textsuperscript{th} (right after pruning), 40K\textsuperscript{th}, 50K\textsuperscript{th}, and 60K\textsuperscript{th} iterations, as shown in Table~\ref{table:mnist_NMF} along with the compression ratio computed as $mn/(k(m+n))$.
Compared with other binary pruning index formats (such as ternary quantization), our proposed binary index factorization yields much higher index compression ratios while maintaining reasonable test accuracy.

\begin{table}[t]
    \caption{(Left): MNIST LeNet-5 accuracy using different rank $k$. At the 20K\textsuperscript{th} iteration, we prune all layers using magnitude-based pruning method \cite{SHan_2015} except FC1 layer where pruning is performed by using Algorithm \ref{alg:nmf}. Retraining is completed at the 60K\textsuperscript{th} iteration. The test accuracy of pre-trained model is 99.2\%. (Right): Index size of FC1 is compared with various index formats. Accuracy is higher than 99.0\% for all formats.}
    \label{table:mnist_NMF}
\begin{center}
%\begin{small}
%\begin{sc}
    \begin{tabular}{c c c c c c}
    \Xhline{2\arrayrulewidth}
    Rank & \multicolumn{4}{c}{Accuracy(\%)} & Comp. \\
    ($k$) & 20K\textsuperscript{th} & 40K\textsuperscript{th} & 50K\textsuperscript{th} & 60K\textsuperscript{th} & Ratio \\
    \Xhline{2\arrayrulewidth}
    4 & 33.69 & 98.96 & 99.03 & 99.07 & 76.9$\times$ \\
    8 & 34.52 & 98.98 & 99.05 & 99.09 & 38.5$\times$ \\
    16 & 30.28 & 99.01 & 99.09 & 99.13 & 19.2$\times$ \\
    32 & 32.52 & 99.01 & 99.08 & 99.13 & 9.6$\times$ \\
    64 & 36.02 & 99.04 & 99.10 & 99.14 & 4.8$\times$ \\
    128 & 34.22 & 99.04 & 99.10 & 99.16 & 2.4$\times$ \\
    256 & 42.86 & 99.07 & 99.14 & 99.19 & 1.2$\times$ \\
    \Xhline{2\arrayrulewidth}
    \end{tabular}
    \hfill
    \begin{tabular}{c c c}
        \Xhline{2\arrayrulewidth}
        Method & Index Size & Comment\\
        \Xhline{2\arrayrulewidth}
        Binary & 50.0KB & 1bit/weight\\
        CSR(16bit) & 45.8KB & \\
        CSR(5bit) & 14.3KB & Relative\\
        Viterbi & 10.0KB & 5X encoder\\
        \hline
        \textbf{Proposed} & \textbf{2.6KB} & $k$=16\\
        \Xhline{2\arrayrulewidth}
    \end{tabular}
%\end{sc}
%\end{small}
\end{center}
\end{table}

%\subsection{Enhancing Sharpness of Masking Layers}

\begin{figure}[t!]
\begin{center}
\includegraphics[width=0.50\linewidth]{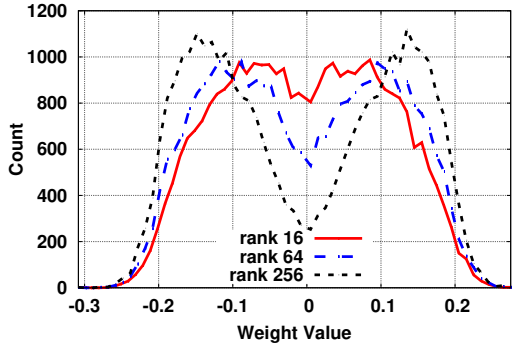}
\caption{Histogram of unpruned weights of FC1 layer at 20K\textsuperscript{th} iteration using LeNet-5 model. Higher $k$ results in more near-zero weights pruned.}
\label{fig:rank_mask_rank}
\end{center}
\end{figure}

In general, the histogram of a weight matrix of a pre-trained model follows a Gaussian distribution \cite{deeplearningbook}.
For magnitude-based pruning, the count of near-zero values significantly reduces after pruning.
To investigate which weights are pruned by different rank, Figure \ref{fig:rank_mask_rank} presents the histogram of unpruned weight values of the FC1 layer of LeNet-5 model at the 20K\textsuperscript{th} iteration right after performing Algorithm \ref{alg:nmf} using the same pruning rate of 95\%.
Results show the count of near-zero weights reduces as rank increases.
Since the total count is the same in Figure \ref{fig:rank_mask_rank} for different rank $k$, a low count of near-zero weights in the histogram implies that fewer weights of large magnitude are pruned.
As such, there is a trade-off between the rank $k$ and accuracy even though accuracy drop is reasonable for a wide range of rank values as shown in Table~\ref{table:mnist_NMF}.

Table \ref{table:mnist_NMF} also compares index size for the FC1 layer using different formats while  keeping pruning rate at 95.0\%.
Reduction in index size for CSR can be achieved using a relative index scheme 
%Amount of index for CSR format can be reduced by using a relative index scheme as
as described in \cite{deepcompression}. % and we assumed 5 bits are used for CSR index.
In the case of Viterbi-based pruning, we assume the Viterbi decompressor has 40 outputs, 10-bit comparators, and a skip state of 1 \cite{lee2018viterbibased}.
%that the number of outputs of Viterbi decompressor is 40 and the number of bits for comparators is 10 when the skip state is 1 \cite{lee2018viterbibased}.
Our proposed pruning method using binary matrix factorization yields the best compression ratio.

\section{Index Matrix Tiling and Weight Manipulation to Lower Rank}

If rank $k$ increases to drop more near-zero weights, compression ratio must be sacrificed.
To decrease the count of near-zero weights after pruning for a fixed rank $k$, we propose two techniques.

\subsection{Tile-Based Binary Matrix Factorization}

\begin{figure*}[t]
	\begin{minipage}{.55\textwidth}
		\includegraphics[width=1\linewidth]{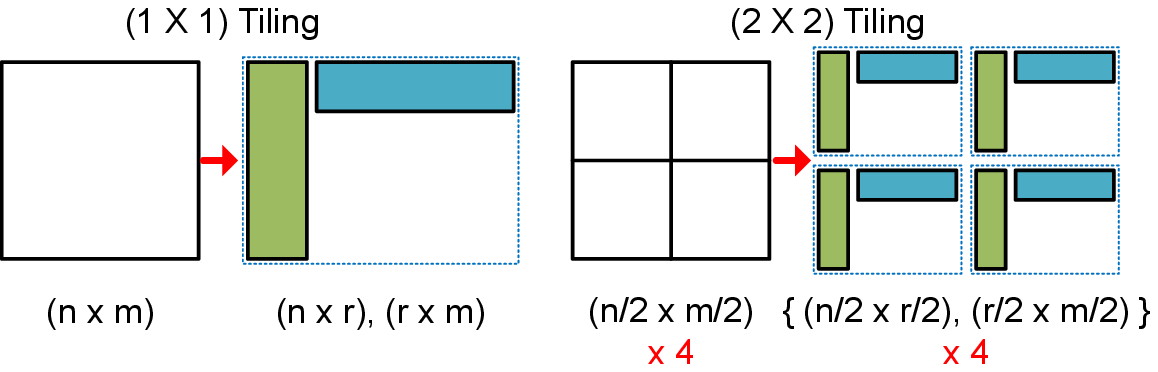}
	\end{minipage}
	\hfill
	\begin{minipage}{.40\textwidth}
		\includegraphics[width=1\linewidth]{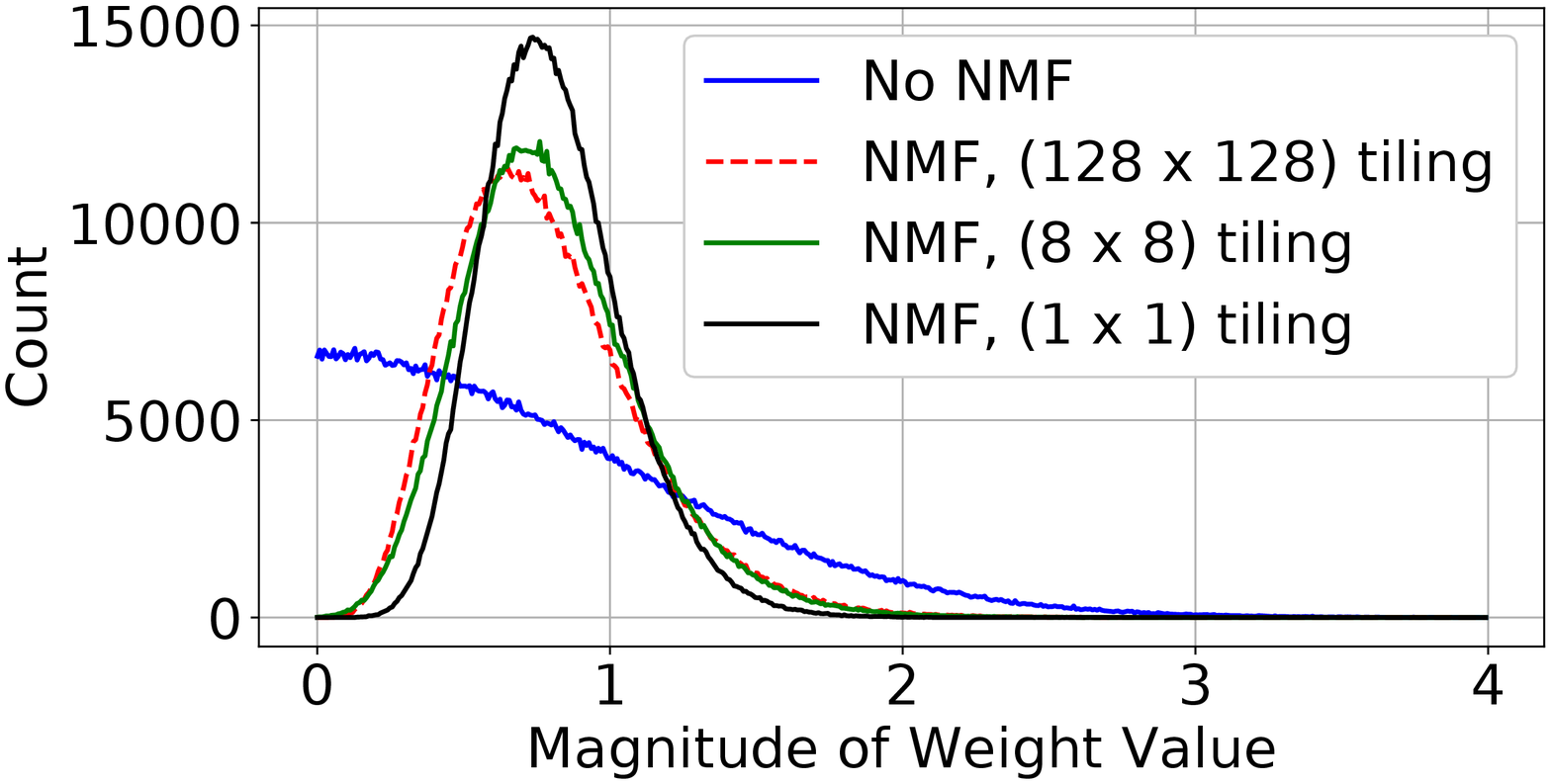}
	\end{minipage}
	\hfill
	\caption{(Left): ($n \times m$) binary matrix factorization using one or four tiles when the compression ratio is the same. (Right): Weight histogram before and after NMF using different number of tiles.}
	\label{fig:nmf_tiling}
\end{figure*}

Weight matrix size increases quadratically with the number of neurons.
If the whole binary matrix multiplication were to be performed within a chip, then on-chip memory size would be prohibitively large.
In order to enhance scalability of our proposed factorization method, we propose tile-based factorization as illustrated in Figure \ref{fig:nmf_tiling}.
A ($n \times m$) binary matrix is first tiled into multiple blocks. Then, each block is factorized independently with the same rank.
Note that tiling size and/or reshaping the original binary index matrix can be varied to match the underlying parallel computing resources.
Because NMF is solved by an iterative method in general, tiling not only reduces the required on-chip memory size but also reduces NMF computation time.

\begin{figure*}[t]
	\begin{minipage}{.45\textwidth}
		\includegraphics[width=1\linewidth]{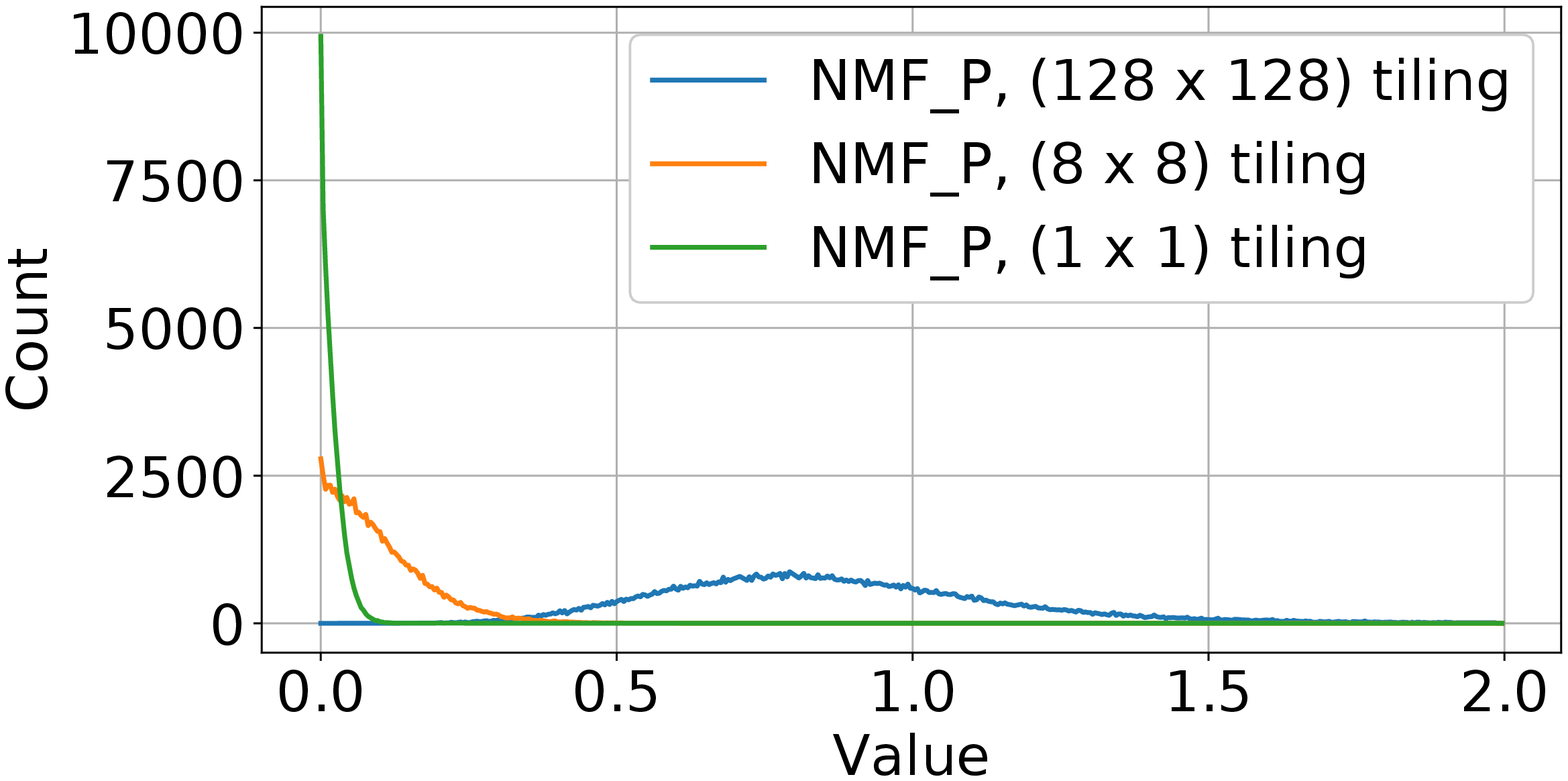}
	\end{minipage}
	\hfill
	\begin{minipage}{.45\textwidth}
		\includegraphics[width=1\linewidth]{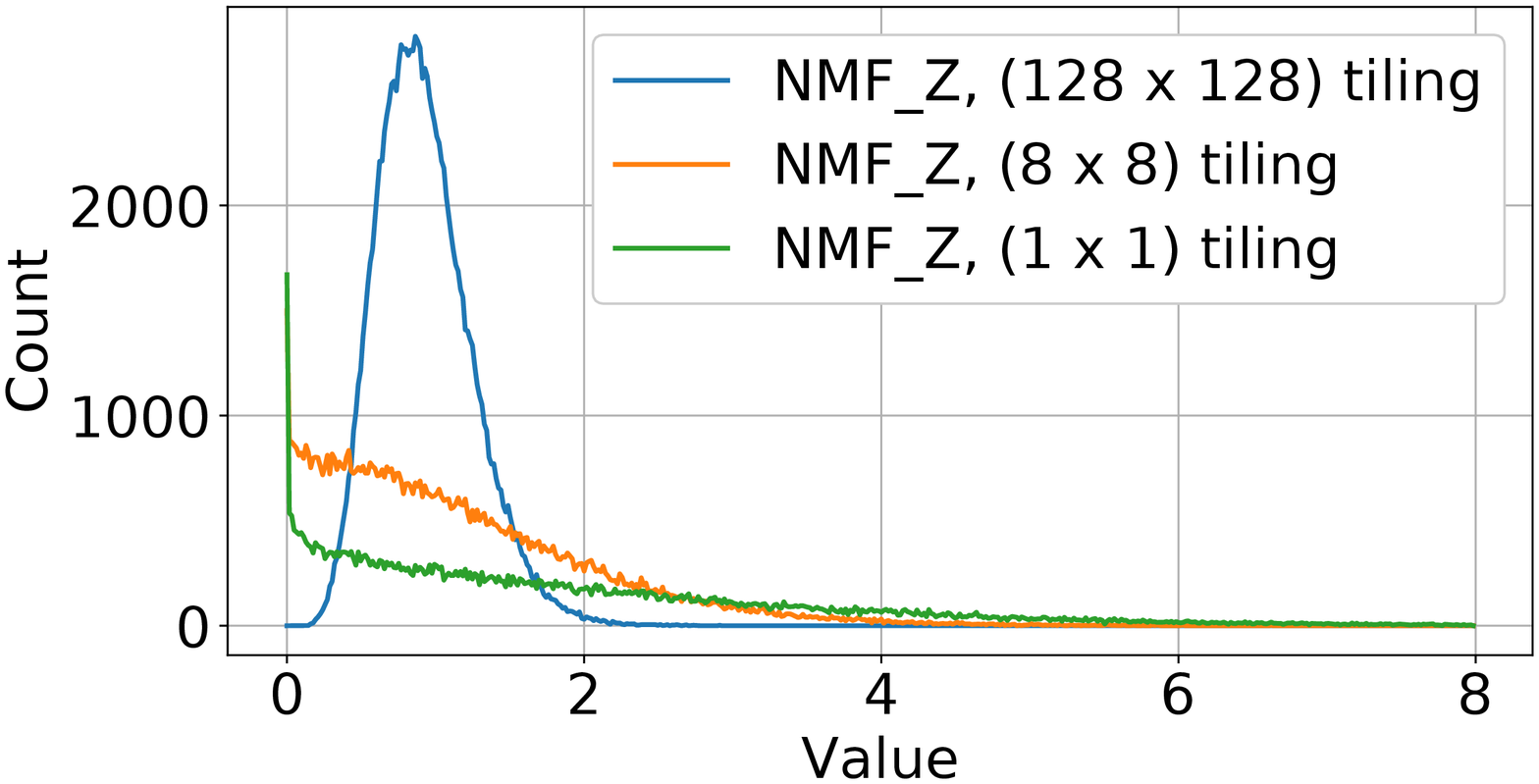}
	\end{minipage}
	\hfill
	\caption{Histogram of $\mM_p$ and $\mM_z$ values using different number of tiles.}
	\label{fig:nmf_tiling2}
\end{figure*}

Tiling also affects the quality of binary matrix factorization.
Note that when $X_1$, $X_2$, $\ldots$, $X_n$ are random samples of size $n$ from a distribution with mean $\mu$ and variance $\sigma^2$, the sample mean $\bar{X}$ shows $E(\bar{X})=\mu$ and $Var(\bar{X})=\sigma^2/n$.
Indeed, Figure \ref{fig:nmf_tiling} depicts large variance of weight values after NMF when the number of tiles increases (a random Gaussian weight distribution is first assumed).
Correspondingly, $\mM_p$ and $\mM_z$ also show larger variance with longer tails in the distribution with more blocks as shown in Figure \ref{fig:nmf_tiling2}.
Such increased variance of $\mM_p$ and $\mM_z$ with smaller sample size is desirable for binary conversion from the NMF result.
For example, compared with \{0.98, 1.0, 1.2\}, \{0.5, 1.0, 1.5\} presents larger spectrum of binary conversion threshold values for $S_p$ and $S_z$ (and hence, increases the chance to further optimize the cost function).

\begin{figure}[t]
\begin{center}
\includegraphics[width=0.65\linewidth]{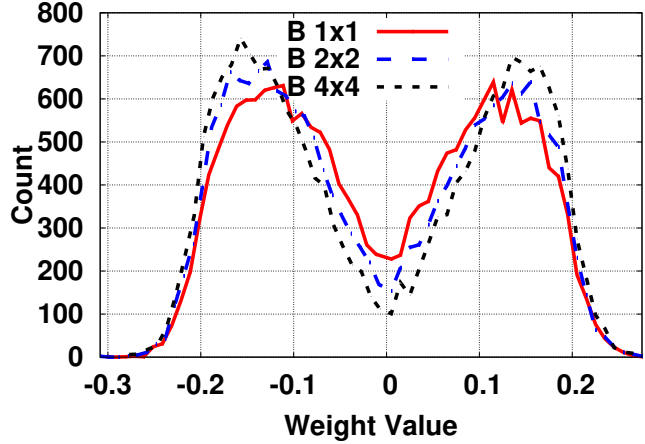}
\caption{Histogram of unpruned FC1 layer weights of LeNet-5 with different block configurations. Each submatrix is factorized by NMF using different rank depending on the tiling plan to present the same overall compression ratio. Rank 128, 64 or 32 is used for ($1\times1$), ($2\times2$), or ($4\times4$) tiling, respectively.}
\label{fig:rank_mask_tile_result}
\end{center}
\end{figure}

Figure \ref{fig:rank_mask_tile_result} plots histograms of unpruned FC1 layer weights of the LeNet-5 model on MNIST across different tiling methods.
FC1 weight matrix (of $500 \times 800$ size) is tiled into 1($1 \times 1$), 4($2 \times 2$), or 16($4 \times 4$) submatrices while the index compression ratio is the same for all three tiling cases.
Since the size of a submatrix differs for each partitioning plan, the rank for factorizing a submatrix is accordingly adjusted in Figure \ref{fig:rank_mask_tile_result}.
Notice that increasing the number of blocks yields deeper drops of near-zero weights.
However, if submatrices have different properties (e.g., an embedding matrix in natural language models) by tiling, each submatrix can optimally choose a different rank.

\subsection{Weight Magnitude Manipulation}
If $Cost$ is given as a magnitude-sum of unintentionally pruned weights, then it is still possible that a large weight can be pruned.
To prevent large weights from being pruned and to keep the definition of $Cost$, the magnitude of weights can be pre-processed.
For example, in the context of magnitude-based pruning, artificially increasing already large weight values can futher lower their probability of being pruned.
Note that such weight magnitude manipulation is only temporarily used for pruning-index data compression, and not for normal training or inference steps.

\begin{figure}[t]
\begin{center}
\includegraphics[width=0.65\linewidth]{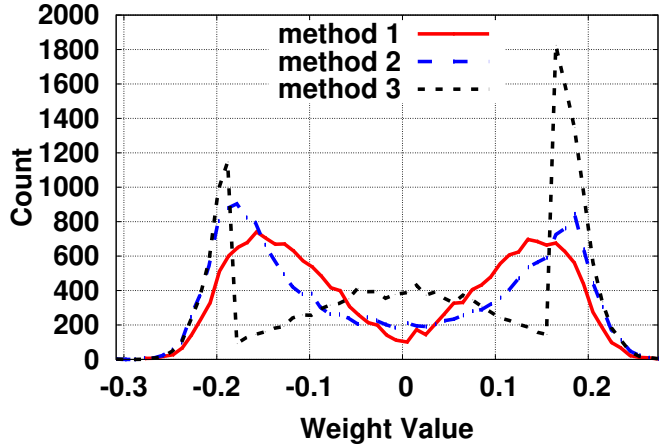}
\caption{Histogram of unpruned FC1 layer weights of LeNet-5 with different weight magnitude manipulation methods (Method 1: no manipulation, Method 2: $\displaystyle \emM_{i,j}$ $\rightarrow$ $\displaystyle \emM^2_{i,j}$, Method 3: $\displaystyle \emM_{i,j}$ is amplified by $1/(1-S)$ if $\displaystyle \emM_{i,j}$ is larger than a threshold value (given by a magnitude-based pruning), where $S$ is the sparsity of $\displaystyle \mW$.}
\label{fig:rank_mask_shape2}
\end{center}
\end{figure}

Figure \ref{fig:rank_mask_shape2} plots histograms of unpruned weights of the FC1 layer of LeNet-5 using different weight-magnitude manipulation methods.
In Method 3, weights larger than a threshold are multiplied by $1/(1-S)$.
We observe the sharpest drop of weights around the threshold value and higher count of large weights in Method 3.
Finding the best weight manipulation method is an open problem.

\section{Experimental Results}

We verify our index data compression technique using ResNet32 \cite{resnet} on CIFAR10, AlexNet \cite{alexnet} on ImageNet, and a recurrent neural network (RNN) model on the PTB dataset \cite{ptb}, as shown in Table \ref{table:final_results}.
We do not apply binary matrix factorization to layers if their parameter size is relatively small.
Such layers usually present small pruning rates \cite{SHan_2015, suyog_prune} and index compression ratio is also small \cite{scalpel2017}.
For example, Viterbi-based index compression techniques yield 1$\times$ compression ratios for the index data of convolutional layers in AlexNet \cite{lee2018viterbibased}.

For ResNet32, our pruning-index compression achieves a 70\% pruning rate with 91.8\% accuracy, providing over 3$\times$ compression as shown in Table \ref{table:final_results} (experimental data using various rank selections id provided in Appendix).
Note that binary pruning-index matrix decomposition is performed after multiplying the weights (exceeding the pruning threshold) by $1/(1-S)$, as a method to drop more near-zero weights.

For AlexNet, we focus on compressing index data of FC5 and FC6 layers occupying $\sim$90\% of the total model size.
Both layers are pruned to a pruning rate of 91\% \cite{SHan_2015} using Algorithm 1.
We achieve over 8$\times$ compression in the pruning-index data while maintaining full-precision accuracy.
FC5 and FC6 weights are tiled into small blocks, given their large matrix size.
BMF of Algorithm~1 is performed on those blocks with ranks 32 and 64, resulting in speed up and reduced $Cost$.

An RNN model with one long-short term memory (LSTM) layer of size 300 \cite{xu2018alternating} on the PTB dataset, with performance measured using Perplexity Per Word (PPW), is pruned by our pruning-index-compression method.
Note that the embedding and softmax matrices usually take a major memory footprint because of increasing vocabulary size in a neural language model while these two matrices have several distinguished properties compared with general weight matrices \cite{GroupReduce}.

To compare the compression ratio of various sparse matrix representation schemes, we choose AlexNet FC5 and FC6 layers and Table \ref{table:alexnet_comp} shows the index size of a binary matrix representation, CSR format with 16-bit indexing, CSR format with 5-bit indexing (relative indexing as introduced in \cite{deepcompression}), Viterbi-based representation \cite{lee2018viterbibased}, and our proposed representation.
Our proposed network pruning algorithm and index representation using binary matrix factorization significantly reduce the amount of indexing data while maintaining sparsity at the same level as fine-grained pruning.
Lastly, while Viterbi-based compression only allows for integer-valued compression ratios, our proposed compression technique enables a much wider range of compression ratios (as rational numbers) by controlling the rank for index pruning.

\newcommand{\mr}[2]{
\multirow{#1}{*}{\begin{tabular}[c]{@{}c@{}}#2\end{tabular}}
}

\begin{table}[t]
\caption{Compression ratio and accuracy of the proposed pruning-index compression method on various DNN models.}
\label{table:final_results}
\vskip 0.15in
\begin{center}\begin{threeparttable}
\begin{tabular}{c|c|c|c|c|c|c}
\Xhline{2\arrayrulewidth}
   \multicolumn{3}{c|}{Pre-trained Model} & \multicolumn{4}{c}{Pruned Model using the Proposed Method} \\ \hline
  Model & Size  & Accuracy & $S$ & Rank & Comp. Ratio & Accuracy \\
\Xhline{2\arrayrulewidth}
       \mr{2}{ResNet32\\on CIFAR10 } & \mr{2}{460.76K} & \mr{2}{92.5\%} & \mr{2}{0.70} & 8/16/32\tnote{1} &  \mr{1}{3.09$\times$} & \mr{1}{91.8\%} \\ \cline{5-7}
       & & & & 8/8/8 & \mr{1}{5.12$\times$} & \mr{1}{91.5}\%  \\
 \hline
      \mr{4}{AlexNet\\on ImageNet\\(FC5, FC6)}    
        & \mr{4}{9K$\times$4K\\(FC5)\\4K$\times$4K\\(FC6)} & \mr{4}{80.3\% (top5)\\57.6\% (top1)} 
        & \mr{4}{0.91} & \mr{2}{32\\\textit{tiled}\tnote{2}} & \mr{2}{8.20$\times$} 
        & \mr{4}{80.4\% (top5)\\57.1\% (top1)\\80.3\% (top5)\\57.2\% (top1)}  \\
       & & & & & & \\  
       \cline{5-7}
       & & & & \mr{2}{64\\\textit{tiled}\tnote{2}} & \mr{2}{4.14$\times$} & \\
       & & & & & & \\
 \hline
     
    \mr{1}{LSTM on PTB} & \mr{1}{6.41M} & \mr{1}{89.6 PPW} & \mr{1}{0.60} & 145 & \mr{1}{1.82$\times$} & 89.0 PPW \\ 
\Xhline{2\arrayrulewidth}
\end{tabular}
\begin{tablenotes}[flushleft]
    \item[1] Different ranks are applied to 3 groups of layers according to the number of input channels (16, 32, or 64).
    \item[2] The FC5 and FC6 layers are tiled to 16$\times$8 blocks (of 576$\times$512 size) and 8$\times$8 blocks (of 512$\times$512 size), respectively.
    %\item[3] Index compression is applied to embedding and softmax layers occupying >90\% of model size.
\end{tablenotes}
\end{threeparttable}\end{center}
\end{table}

\begin{table}[t]
    \caption{Index size of FC5 and FC6 layers of AlexNet is compared with various index formats when $S$ is to be the same as 0.91 for both FC5 and FC6. Top-1 accuracy is higher than 57.0\% for all formats.}
    \label{table:alexnet_comp}
\begin{center}
%\begin{small}
%\begin{sc}
    \begin{tabular}{c c c c c}
        \Xhline{2\arrayrulewidth}
        Method & FC5 Index Size & FC6 Index Size & Sum & Comment\\
        \hline
        Binary & 4608KB & 2048KB & 6656KB & 1bit/weight\\
        CSR(16bit) & 6962KB & 3099KB & 10061KB & \\
        CSR(5bit) & 2176KB & 968KB & 3144KB & Relative Indexing \\
        Viterbi & 922KB & 410KB & 1331KB & 5X Encoder\\
        \hline
        \textbf{Proposed} & \textbf{556KB} & \textbf{256KB} & \textbf{812KB} & $k$=32, tiled\\
        \Xhline{2\arrayrulewidth}
    \end{tabular}
%\end{sc}
%\end{small}
\end{center}
\end{table}

\section{Conclusion}

This paper proposes a fine-grained network pruning technique to produce low-rank binary index matrices.
We confirm various DNN models can be pruned by our low-rank indexing scheme while preserving sparsity.
Such binary matrix multiplication enables not only high compression ratios, but highly parallel operations of sparse networks.
Our proposed tiling method and weight magnitude manipulation schemes further lower rank.
Unlike previous studies, this work demonstrates that fine-grained pruning can be represented by a highly regular structured format.
%
%\subsubsection*{Acknowledgments}

%Use unnumbered third level headings for the acknowledgments. All acknowledgments
%go at the end of the paper. Do not include acknowledgments in the anonymized
%submission, only in the final paper.

%\section*{References}

\small
\bibliography{neurips_2019_index}

\newpage

\appendix

\section{Supplementary Experimental Results on ResNet32 using CIFAR-10}
Table \ref{table:resnet_nmf} lists the accuracy of ResNet32 using various ranks and pruning rates. The pruned models using our proposed binary matrix decomposition show slightly degraded accuracy compared to the baseline pruning method (the bottom row of Table \ref{table:resnet_nmf}). It can be observed that in general, there exists a trade-off between the ranks and compression ratios.

\begin{table}[h]
\caption{Model accuracy of ResNet32 with CIFAR-10 using different ranks and pruning rates. The bottom row presents the results of baseline pruning method without binary matrix factorization.}
\label{table:resnet_nmf}
\vskip 0.15in
\begin{center}\begin{threeparttable}%\begin{sc}
\begin{tabular}{c|c|ccc}
\toprule
 & & \multicolumn{3}{c}{Pruning Rate ($S$)} \\
 \cline{3-5}
Rank\tnote{1} & Comp. Ratio & 0.60 & 0.70 & 0.80 \\
\midrule
4/4/4 & 10.29$\times$ & 92.0\% & 91.5\% & 90.5\%\\ 
4/8/16 & 6.74$\times$ & 92.2\% & 91.5\% & 90.9\%\\
8/8/8 &  5.12$\times$ & 92.0\% & 91.5\% & 90.9\%\\
8/16/32 & 3.09$\times$ & 92.4\% & 91.8\% & 91.1\%\\
16/16/16 & 2.56$\times$ & 92.2\% & 91.8\% & 91.0\%\\
16/32/64 & 1.55$\times$ & 92.3\% & 91.7\% & 91.2\%\\
w/o BMF & 1$\times$ & 92.4\% & 92.2\% & 92.0\%\\
\bottomrule
\end{tabular}
%\end{sc}
\begin{tablenotes}
    \item[1] Different ranks are applied to 3 groups of layers according to the number of input channels (16, 32, or 64).
    \item[2] Retraining is performed for 65K iterations.
\end{tablenotes}
\end{threeparttable}
\end{center}
\vskip -0.1in
\end{table}

\end{document}

%% file: math_commands.tex
\usepackage{amsmath,amsfonts,bm}

\def\eqref#1{equation~\ref{#1}}
\def\1{\bm{1}}

\def\mH{{\bm{H}}}
\def\mI{{\bm{I}}}

\def\mM{{\bm{M}}}

\def\mW{{\bm{W}}}

\DeclareMathAlphabet{\mathsfit}{\encodingdefault}{\sfdefault}{m}{sl}
\SetMathAlphabet{\mathsfit}{bold}{\encodingdefault}{\sfdefault}{bx}{n}

\def\emI{{I}}

\def\emM{{M}}

\def\emW{{W}}

\newcommand{\R}{\mathbb{R}}

%% file: neurips_2019_index.bbl
\begin{thebibliography}{27}
\providecommand{\natexlab}[1]{#1}
\providecommand{\url}[1]{\texttt{#1}}
\expandafter\ifx\csname urlstyle\endcsname\relax
  \providecommand{\doi}[1]{doi: #1}\else
  \providecommand{\doi}{doi: \begingroup \urlstyle{rm}\Url}\fi

\bibitem[Ahn et~al.(2019)Ahn, Lee, Kim, and Kim]{viterbi_quantized}
D.~Ahn, D.~Lee, T.~Kim, and J.-J. Kim.
\newblock Double {Viterbi}: Weight encoding for high compression ratio and fast
  on-chip reconstruction for deep neural network.
\newblock In \emph{International Conference on Learning Representations
  (ICLR)}, 2019.

\bibitem[Chen et~al.(2018)Chen, Si, Li, Chelba, and Hsieh]{GroupReduce}
P.~Chen, S.~Si, Y.~Li, C.~Chelba, and C.-J. Hsieh.
\newblock {GroupReduce}: Block-wise low-rank approximation for neural language
  model shrinking.
\newblock In \emph{Advances in Neural Information Processing Systems}, 2018.

\bibitem[Dean et~al.(2012)Dean, Corrado, Monga, Chen, Devin, Mao, aurelio
  Ranzato, Senior, Tucker, Yang, Le, and Ng]{large_scale_distributed}
J.~Dean, G.~Corrado, R.~Monga, K.~Chen, M.~Devin, M.~Mao, M.~aurelio Ranzato,
  A.~Senior, P.~Tucker, K.~Yang, Q.~V. Le, and A.~Y. Ng.
\newblock Large scale distributed deep networks.
\newblock In \emph{Advances in Neural Information Processing Systems}, pages
  1223--1231, 2012.

\bibitem[Frankle and Carbin(2019)]{lottery}
J.~Frankle and M.~Carbin.
\newblock The lottery ticket hypothesis: Finding sparse, trainable neural
  networks.
\newblock In \emph{International Conference on Learning Representations
  (ICLR)}, 2019.

\bibitem[Goodfellow et~al.(2016)Goodfellow, Bengio, and
  Courville]{deeplearningbook}
I.~Goodfellow, Y.~Bengio, and A.~Courville.
\newblock \emph{Deep Learning}.
\newblock MIT Press, 2016.
\newblock \url{http://www.deeplearningbook.org}.

\bibitem[Guo et~al.(2016)Guo, Yao, and Chen]{DNS}
Y.~Guo, A.~Yao, and Y.~Chen.
\newblock Dynamic network surgery for efficient {DNNs}.
\newblock In \emph{Advances in Neural Information Processing Systems}, 2016.

\bibitem[Han et~al.(2015)Han, Pool, Tran, and Dally]{SHan_2015}
S.~Han, J.~Pool, J.~Tran, and W.~J. Dally.
\newblock Learning both weights and connections for efficient neural networks.
\newblock In \emph{Advances in Neural Information Processing Systems}, pages
  1135--1143, 2015.

\bibitem[Han et~al.(2016{\natexlab{a}})Han, Liu, Mao, Pu, Pedram, Horowitz, and
  Dally]{EIE}
S.~Han, X.~Liu, H.~Mao, J.~Pu, A.~Pedram, M.~A. Horowitz, and W.~J. Dally.
\newblock {EIE:} efficient inference engine on compressed deep neural network.
\newblock In \emph{Proceedings of the 43rd International Symposium on Computer
  Architecture}, pages 243--254, 2016{\natexlab{a}}.

\bibitem[Han et~al.(2016{\natexlab{b}})Han, Mao, and Dally]{deepcompression}
S.~Han, H.~Mao, and W.~J. Dally.
\newblock Deep compression: Compressing deep neural networks with pruning,
  trained quantization and {Huffman} coding.
\newblock In \emph{International Conference on Learning Representations
  (ICLR)}, 2016{\natexlab{b}}.

\bibitem[He et~al.(2016)He, Zhang, Ren, and Sun]{resnet}
K.~He, X.~Zhang, S.~Ren, and J.~Sun.
\newblock Deep residual learning for image recognition.
\newblock \emph{2016 IEEE Conference on Computer Vision and Pattern Recognition
  (CVPR)}, pages 770--778, 2016.

\bibitem[He et~al.(2017)He, Zhang, and Sun]{channel_pruning_1}
Y.~He, X.~Zhang, and J.~Sun.
\newblock Channel pruning for accelerating very deep neural networks.
\newblock In \emph{The IEEE International Conference on Computer Vision
  (ICCV)}, Oct 2017.

\bibitem[Krizhevsky et~al.(2012)Krizhevsky, Sutskever, and Hinton]{alexnet}
A.~Krizhevsky, I.~Sutskever, and G.~E. Hinton.
\newblock Imagenet classification with deep convolutional neural networks.
\newblock In F.~Pereira, C.~J.~C. Burges, L.~Bottou, and K.~Q. Weinberger,
  editors, \emph{Advances in Neural Information Processing Systems 25}, pages
  1097--1105. Curran Associates, Inc., 2012.

\bibitem[LeCun et~al.(1990)LeCun, Denker, and Solla]{optimalbrain}
Y.~LeCun, J.~S. Denker, and S.~A. Solla.
\newblock Optimal brain damage.
\newblock In \emph{Advances in Neural Information Processing Systems}, pages
  598--605, 1990.

\bibitem[Lee et~al.(2018{\natexlab{a}})Lee, Ahn, Kim, Chuang, and
  Kim]{lee2018viterbibased}
D.~Lee, D.~Ahn, T.~Kim, P.~I. Chuang, and J.-J. Kim.
\newblock Viterbi-based pruning for sparse matrix with fixed and high index
  compression ratio.
\newblock In \emph{International Conference on Learning Representations
  (ICLR)}, 2018{\natexlab{a}}.

\bibitem[Lee et~al.(2018{\natexlab{b}})Lee, Kapoor, and Kim]{DeepTwist}
D.~Lee, P.~Kapoor, and B.~Kim.
\newblock Deeptwist: Learning model compression via occasional weight
  distortion.
\newblock \emph{arXiv:1810.12823}, 2018{\natexlab{b}}.

\bibitem[Lee and Seung(1999)]{NMF}
D.~D. Lee and H.~S. Seung.
\newblock Learning the parts of objects by non-negative matrix factorization.
\newblock \emph{Nature}, 401\penalty0 (6755):\penalty0 788--791, 1999.

\bibitem[Liu et~al.(2019)Liu, Sun, Zhou, Huang, and Darrell]{rethinking}
Z.~Liu, M.~Sun, T.~Zhou, G.~Huang, and T.~Darrell.
\newblock Rethinking the value of network pruning.
\newblock In \emph{International Conference on Learning Representations
  (ICLR)}, 2019.

\bibitem[Marcus et~al.(1994)Marcus, Kim, Marcinkiewicz, MacIntyre, Bies,
  Ferguson, Katz, and Schasberger]{ptb}
M.~Marcus, G.~Kim, M.~A. Marcinkiewicz, R.~MacIntyre, A.~Bies, M.~Ferguson,
  K.~Katz, and B.~Schasberger.
\newblock The penn treebank: Annotating predicate argument structure.
\newblock In \emph{Proceedings of the Workshop on Human Language Technology},
  pages 114--119, 1994.

\bibitem[Molchanov et~al.(2017)Molchanov, Ashukha, and Vetrov]{sparseVD}
D.~Molchanov, A.~Ashukha, and D.~P. Vetrov.
\newblock Variational dropout sparsifies deep neural networks.
\newblock In \emph{International Conference on Machine Learning ({ICML})},
  pages 2498--2507, 2017.

\bibitem[Narang et~al.(2017)Narang, Undersander, and Diamos]{block-sparse}
S.~Narang, E.~Undersander, and G.~F. Diamos.
\newblock Block-sparse recurrent neural networks.
\newblock \emph{arXiv:1711.02782}, 2017.

\bibitem[Wen et~al.(2016)Wen, Wu, Wang, Chen, and Li]{channel_pruning_2}
W.~Wen, C.~Wu, Y.~Wang, Y.~Chen, and H.~Li.
\newblock Learning structured sparsity in deep neural networks.
\newblock In \emph{Advances in Neural Information Processing Systems}, pages
  2082--2090, 2016.

\bibitem[Xu et~al.(2018)Xu, Yao, Lin, Ou, Cao, Wang, and
  Zha]{xu2018alternating}
C.~Xu, J.~Yao, Z.~Lin, W.~Ou, Y.~Cao, Z.~Wang, and H.~Zha.
\newblock Alternating multi-bit quantization for recurrent neural networks.
\newblock In \emph{International Conference on Learning Representations
  (ICLR)}, 2018.

\bibitem[Yu et~al.(2017)Yu, Lukefahr, Palframan, Dasika, Das, and
  Mahlke]{scalpel2017}
J.~Yu, A.~Lukefahr, D.~Palframan, G.~Dasika, R.~Das, and S.~Mahlke.
\newblock Scalpel: Customizing {DNN} pruning to the underlying hardware
  parallelism.
\newblock In \emph{Proceedings of the 44th Annual International Symposium on
  Computer Architecture}, pages 548--560, 2017.

\bibitem[Zhang et~al.(2016)Zhang, Du, Zhang, Lan, Liu, Li, Guo, Chen, and
  Chen]{cambricon-x}
S.~Zhang, Z.~Du, L.~Zhang, H.~Lan, S.~Liu, L.~Li, Q.~Guo, T.~Chen, and Y.~Chen.
\newblock Cambricon-x: An accelerator for sparse neural networks.
\newblock In \emph{The 49th Annual IEEE/ACM International Symposium on
  Microarchitecture}, pages 20:1--20:12, 2016.

\bibitem[Zhang et~al.(2007)Zhang, Li, Ding, and Zhang]{BMF}
Z.~Zhang, T.~Li, C.~H.~Q. Ding, and X.~Zhang.
\newblock Binary matrix factorization with applications.
\newblock In \emph{IEEE International Conference on Data Mining}, 2007.

\bibitem[Zhu and Gupta(2017)]{suyog_prune}
M.~Zhu and S.~Gupta.
\newblock To prune, or not to prune: exploring the efficacy of pruning for
  model compression.
\newblock \emph{CoRR}, abs/1710.01878, 2017.

\bibitem[Zitnik and Zupan(2012)]{Zitnik2012}
M.~Zitnik and B.~Zupan.
\newblock Nimfa: A python library for nonnegative matrix factorization.
\newblock \emph{Journal of Machine Learning Research}, 13:\penalty0 849--853,
  2012.

\end{thebibliography}
